\documentclass[letterpaper]{article}
\usepackage{spconf,amsmath,graphicx}
\usepackage{hyperref} 
\usepackage{booktabs, multicol, multirow} 
\usepackage{color,soul} 
\usepackage[colorinlistoftodos]{todonotes} 
\usepackage{rotating} 
\usepackage{tabularx} 

\def\Nclasses{{5 }}
\def\Nimages{4,298 }
\def\Npatients{1,675 }
\def\errorbefore{9.6 } 
\def\errorafter{5.9 } 
\def\aucbefore{0.994 } 
\def\aucafter{0.998 } 
\title{Anatomy-specific classification of medical images using deep convolutional nets}
\name{\small Holger R. Roth, Christopher T. Lee, Hoo-Chang Shin, Ari Seff, Lauren Kim, Jianhua Yao, Le Lu, Ronald M. Summers\thanks{This work was supported by the Intramural Research Program of the NIH Clinical Center. Download the software/code from this paper \href{https://github.com/rsummers11/CADLab/tree/master/CNNSliceClassifier}{here}. Contact: \href{mailto:holger.roth@nih.gov}{holger.roth@nih.gov} or \href{mailto:rms@nih.gov}{rms@nih.gov}}}
\address{\small Imaging Biomarkers and Computer-Aided Diagnosis Laboratory\\
\small Radiology and Imaging Sciences Department\\
\small National Institutes of Health Clinical Center\\
\small Bethesda, MD 20892-1182, USA}
\begin{document}
\maketitle
\begin{abstract}
Automated classification of human anatomy is an important prerequisite for many computer-aided diagnosis systems. The spatial complexity and variability of anatomy throughout the human body makes classification difficult. ``Deep learning'' methods such as convolutional networks (ConvNets) outperform other state-of-the-art methods in image classification tasks. In this work, we present a method for organ- or body-part-specific anatomical classification of medical images acquired using computed tomography (CT) with ConvNets. We train a ConvNet, using \Nimages separate axial 2D key-images to learn \Nclasses anatomical classes. Key-images were mined from a hospital PACS archive, using a set of \Npatients patients. We show that a data augmentation approach can help to enrich the data set and improve classification performance. Using ConvNets and data augmentation, we achieve anatomy-specific classification error of \errorafter\% and area-under-the-curve (AUC) values of an average of \aucafter in testing. We demonstrate that deep learning can be used to train very reliable and accurate classifiers that could initialize further computer-aided diagnosis.
\end{abstract}
\section{INTRODUCTION}
\label{sec:intro}
\begin{figure}[htb!]
	\centering
		\includegraphics[width=7.5cm]{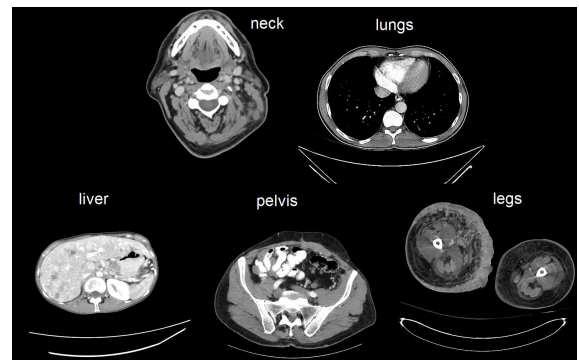}
	\caption{Example key-images of \Nclasses classes of anatomy in our data set: neck, lungs, liver, pelvis and legs.}
\label{fig:organs}
\end{figure}

Medical image classification can be an important component of many computer aided detection (CADe) and diagnosis (CADx) systems. Achieving high accuracies for automated classification of anatomy is a challenging task, given the vast scope of anatomic variation. In this work, our aim is to automatically classify axial CT images into \Nclasses anatomical classes (see Fig. \ref{fig:organs}). This aim is achieved by mining radiological reports that refer to key-images and associated DICOM image tags manually in order to establish a ground truth for training and testing. Using computer vision and medical image computing techniques, we were able to train the computer to replicate these classes with low error rates.
\section{METHOD}
Recently, the availability of large annotated training sets and the accessibility of affordable parallel computing resources via GPUs have made it feasible to train ``deep'' convolutional networks (ConvNets). ConvNets have popularized the topic of ``deep learning'' in computer vision research \cite{jones2014computer}. Through the use of ConvNets, not only have great advances been made in the classification of natural images \cite{krizhevsky2012imagenet}, but substantial advancements have also been made in biomedical applications, such as digital pathology \cite{cirecsan2013mitosis}. Additionally, recent work has shown how the implementation of ConvNets can substantially improve the performance of state-of-the-art CADe systems \cite{Prasoon2013deep, roth2014new, roth2014detection, li2014medical}.
\subsection{Convolutional networks}
\label{sec:convnet}
\begin{figure}[htb!]
	\centering
		\includegraphics[width=8.0cm]{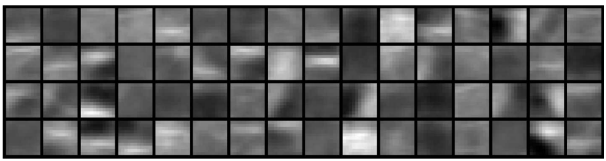}
	\caption{The first layer of learned convolutional kernels of a ConvNet trained on medical CT images.}
	\label{fig:conv1}
\end{figure}

\begin{figure}[htb!]
	\centering
		\includegraphics[width=8.5cm]{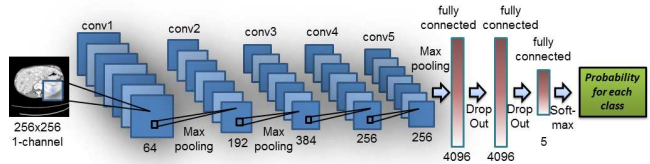}
	\caption{ConvNet applied to an axial CT image. The number of convolutional filters and neural network connections for each layer are as shown.}
	\label{fig:convnet}
\end{figure}

In this work, we apply ConvNets to build an anatomy-specific classifier for CT images. ConvNets are named for their convolutional filters which are used to compute image features for classification. In this work, we use 5 cascaded layers of convolutional filters. All convolutional filter kernel elements are trained from the data in a supervised fashion. This has major advantages over more traditional CAD approaches that use hand-crafted features, designed from human experience. This means that ConvNets have a better chance of capturing the ``essence'' of the imaging data set used for training than when using hand-crafted features \cite{jones2014computer}. Examples of trained filters of the first convolutional layer can be seen in Fig. \ref{fig:conv1}. These first-layer filters capture low spatial frequency signals. In contrast, a mixed set of low and high frequency patterns exists in the first convolutional layer shown in \cite{roth2014new, roth2014detection}. This indicates that the essential information of this task of classifying holistic slice-based body regions lies in the low frequency spatial intensity contrasts. These automatically learned low frequency filters need no tuning by hand, which is different from using intensity histograms, e.g. \cite{feulner2009estimating,dicken2010rapid}.
In-between convolutional layers, the ConvNet performs \textit{max-pooling} operations in order to summarize feature responses across non-overlapping neighboring pixels (see Fig. \ref{fig:convnet}). This allows the ConvNet to learn features that are invariant to spatial variations of objects in the images. Feature responses after the 5th convolutional layer feed into a \textit{fully-connected} neural network. This network learns how to interpret the feature responses and make anatomy-specific classifications. Our ConvNet uses a final \textit{softmax} layer which provides a probability for each object class (see Fig. \ref{fig:convnet}). In order to avoid overfitting, the fully-connected layers are constrained, using the \textit{``DropOut''} method \cite{srivastava2014dropout}. \textit{DropOut} behaves as a regularizer when training the ConvNet by preventing co-adaptation of units in the neural network. We use an open-source implementation (\textit{cuda-convnet2}\footnote{\url{https://code.google.com/p/cuda-convnet2}}) by Krizhevsky et al. \cite{krizhevsky2012imagenet,krizhevsky2014one} which efficiently trains the ConvNet, using GPU acceleration. Further speed-ups are achieved using rectified linear units as neuron activation function instead of the traditional neuron model $f(x) = \tanh(x)$ or $f(x) = (1 + e^{-x})^{-1}$ in both training and evaluation \cite{krizhevsky2012imagenet}. 
\subsection{Data mining of key-images}
We retrieve medical images (many related to liver disease) from the Picture Archiving and Communication System (PACS) of the Clinical Center of the National Institutes of Health by searching for a set of keywords in the radiological reports. Then, each image is assigned a ground truth label based on the \textit{`StudyDescription'} and \textit{`BodyPartExamined'} DICOM tags (manually corrected if necessary). This results in \Nclasses classes of images as shown in Fig. \ref{fig:organs}. Images which show anatomies of multiple classes at once are duplicated and each image copy is assigned one of the class labels. This case commonly occurs at the transition region between lung and liver. Our ConvNet assigns equal probabilities for each class in these regions.
\subsection{Data augmentation}
\label{sec:augmentation}
\begin{figure}[htb!]
	\centering
		\includegraphics[width=7.7cm]{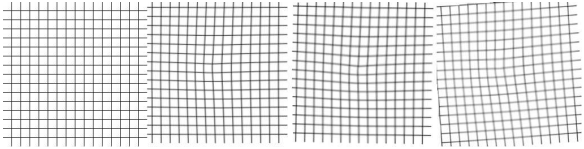}
	\caption{Data augmentation using varying random transformations, rotations and non-rigid deformations using thin-plate-spline (TPS) interpolations on an example image grid.}
	\label{fig:tps_deform}
\end{figure}

We enrich our data set by applying spatial deformations to each image, using random translation, rotations and non-rigid deformations. Each non-rigid training deformation $t$ is computed by fitting a thin-plate-spline (TPS) to a regular grid of 2D control points $\left\{\omega_i;i=1,2,…,K\right\}$. These control points can be randomly transformed at the 2D slice level and a deformed image can be generated using a radial basis function $\phi(r)$:
\begin{equation}
	t(x) = \sum^K_{i=1} c_i \phi\left(\left\|x-\omega_i\right\|\right).
\end{equation}
We use $\phi(r)=r^2log⁡(r)$ which is commonly applied for TPS. A typical TPS deformation field and deformed variations of an example image grid are shown in Fig. \ref{fig:tps_deform}. The variation of translation $t$, rotation $r$ and non-rigid deformations $d$ are a useful way to increase the variety and sample space of available training data, resulting in $N_{\mathrm{aug.}} = N\times N_t\times N_r \times N_d$ variations of the imaging data. The maximum amounts of translation, rotation and non-rigid deformation are chosen such that the resulting deformations resemble plausible physical variations of the medical images. This approach is commonly referred to as data augmentation and can help avoid overfitting \cite{krizhevsky2012imagenet}. Our set of $N_{\mathrm{aug.}}$ axial images are then rescaled to $256 \times 256$ and used to train a ConvNet with a standard architecture for multi-class image classification (as described in Sec. \ref{sec:convnet}).
\section{RESULTS}
\subsection{Key-image data set}
\begin{figure}[htb!]
	\centering
		\includegraphics[width=8.5cm]{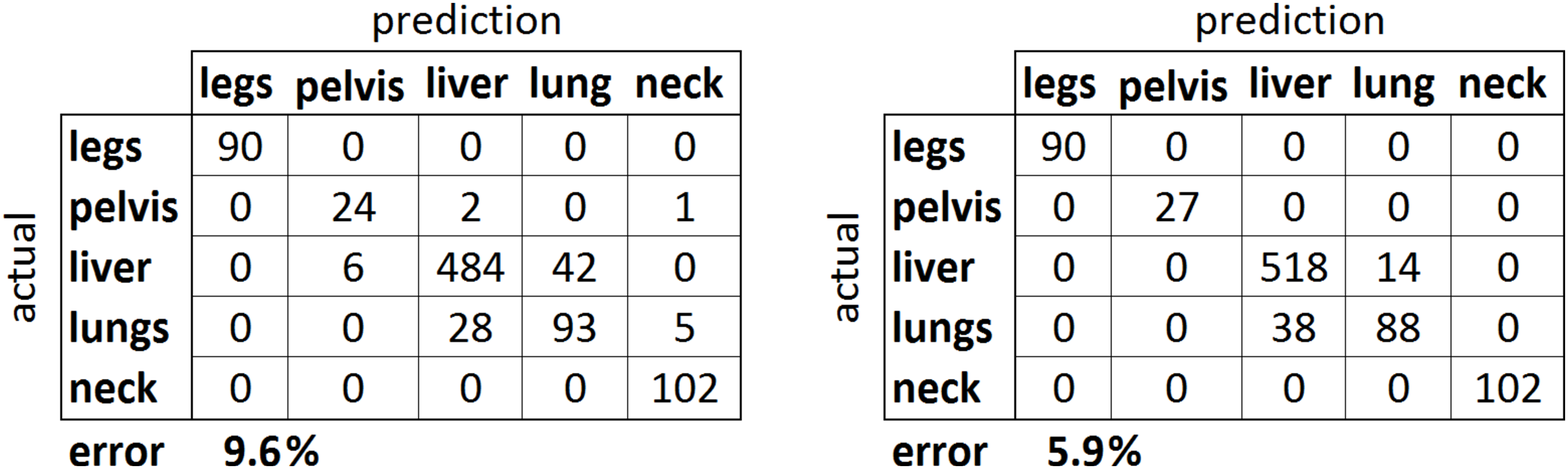}
	\caption{Confusion matrices on the original test images before$^1$ and after$^2$ data augmentation.}
	\label{fig:confusion_matrix}
\end{figure}

\begin{figure}[htb!]
	\centering
		\includegraphics[width=5.0cm]{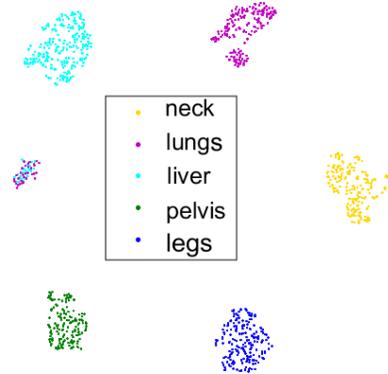}
	\caption{2D embedding of ConvNet features using t-SNE on a subset of test images. Each dot represents a key-image in feature space. The color-coding is based on the ground truth label for each key-image.}
	\label{fig:tsne}
\end{figure}

We use 80 \% of our total dataset for training a multi-class ConvNet as described in Sec. \ref{sec:convnet}. and reserve 20 \% for testing purposes. Our data augmentation step (see Sec \ref{sec:augmentation}) increases the amount of training and testing data drastically, as shown in Table \ref{tab:results}. The number of deformations for each anatomical class is chosen so that the resulting augmented images build a more balanced and enriched data set. We use $N_t=2$ and $N_r=2$ while adjusting $N_d$ for each class to achieve a balanced data set. Table \ref{tab:results} further shows that data augmentation helps to reduce classification errors from \errorbefore\% to \errorafter\% in testing and furthermore improve the average area-under-the-curve (AUC) values from \aucbefore to \aucafter using receiver-operating-characteristic (ROC) analysis. Confusion matrices shown in Fig. \ref{fig:confusion_matrix} show a clear reduction of mis-classification after using data augmentation when testing on the original test set. We further illustrate the feature space of our trained ConvNet using t-SNE \cite{van2008visualizing,donahue2013decaf} in Fig. \ref{fig:tsne}. A clear separation of most classes can be observed. An overlapping cluster can be seen at the interface between the lungs and liver images. This is caused by key-images that show both lungs and livers being near the diaphragm region. 
\begin{table}[htbp]
	\label{tab:results}
  \centering
  \caption{Image data set before$^1$ and after$^2$ data augmentation. An improvement of both error rate and AUC values can be achieved by using data augmentation.}
    \begin{tabular}{rrrrr}
		\toprule
    \toprule
    \textbf{Organ} & \textbf{\#$^1$} & \textbf{\#$^2$} & \textbf{AUC$^1$} & \textbf{AUC$^2$} \\
    \midrule
    \textbf{leg}    & 477  & 24,804 & 1.000 & 1.000 \\
    \textbf{pelvis} & 104  & 22,048 & 0.996 & 1.000 \\
    \textbf{liver}  & 2,684 & 32,208 & 0.994 & 0.999 \\
    \textbf{lung}   & 590  & 25,960 & 0.981 & 0.999 \\
    \textbf{neck}   & 443  & 23,036 & 0.999 & 1.000 \\
		\bottomrule
    \textit{Sum/Mean AUC} & \textit{\Nimages} & \textit{12,8056} & \textbf{\aucbefore} & \textbf{\aucafter} \\
    \textbf{Error} & \textbf{9.6\%} & \textbf{5.9\%} &  &  \\
    \bottomrule
		\bottomrule
    \end{tabular}%
  \label{Image data set before$^1$ and after$^2$ data augmentation.}%
\end{table}%
\subsection{Full torso CT volume}
\label{sec:torso_ct}
\begin{figure}[htb!]
	\centering
		\includegraphics[width=8.5cm]{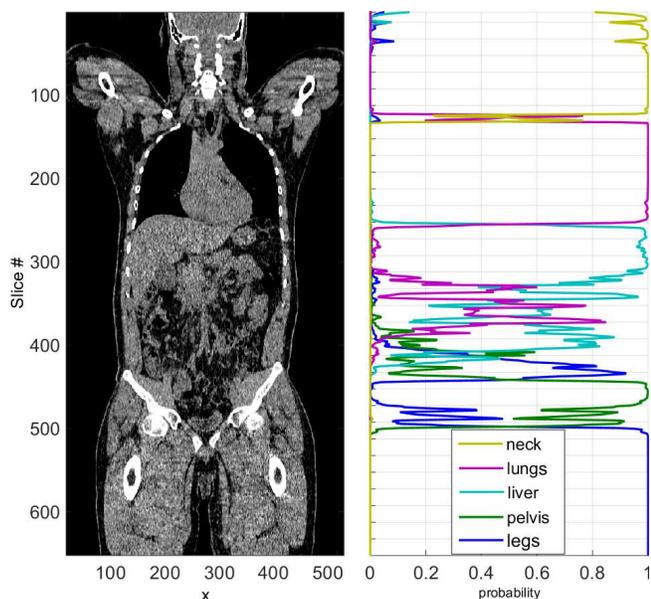}
	\caption{Organ-specific probabilities for a whole-body CT scan.}
	\label{fig:torso_ct}
\end{figure}

For qualitative evaluation, we also apply our trained ConvNet classifier on a full torso CT examination on a slice-by-slice basis (dimensions of $[512, 512, 652]$ and $[0.98, 0.98, 1.5]$ mm voxel spacing). The resulting anatomy-specific probabilities for each slice are plotted as profiles next to the coronal slice of the CT volume in Fig. \ref{fig:torso_ct}. Note how the interface between the lungs and liver at the level of the diaphragm is captured by roughly equal probabilities of the ConvNet. This classification result is achieved in less than 1 minute on a modern desktop computer and GPU card (Dell Precision T7500, 24GB RAM, NVIDIA Titan Z). 
\section{DISCUSSION}
\label{sec:conclusions}
This work demonstrates how deep ConvNets can be applied to effective anatomy-specific classification of medical images. Similar motives to ours are explored in content-based image retrieval methods \cite{akgul2011content}. However, association based on clinical reports and image scans can be very loose. This makes retrieval based on clinical reports difficult. In this paper, we focus on manually labeled key-images that allow us to train an anatomy-specific classifier. Other related work includes the \textit{ImageCLEF} medical image annotation tasks of 2005-2007. However, these tasks used highly subsampled 2D version of medical images ($32 \times 32$ pixels) \cite{muller2007overview}. Methods applied to the \textit{ImageCLEF} tasks included using local image descriptors and intensity histograms in a bag-of-features approach \cite{deselaers2008deformations}. We concentrate on classifying images much closer to their original $512 \times 512$ resolution, namely rescaled to $256 \times 256$. We show that ConvNets can model this higher detail in the images and generalize well to large variations found in medical imaging data with promising quantitative and qualitative results. Some axial slices in the lower abdomen had erroneously high probabilities for lung or legs. Here, it could be beneficial to introduce an additional class of `lower abdomen'. Our method could be easily extended to include further augmentation such as image scales in order to model variations in patient sizes. This type of anatomy classifier could be employed as an initialization step for further and more detailed analysis, such as disease and organ specific computer-aided detection and/or diagnosis.
\bibliographystyle{ieeetr} 
\bibliography{references_isbi}
\end{document}